\documentclass{article}  

\usepackage[preprint]{automl}

\usepackage{microtype} 
\usepackage{booktabs}  
\usepackage{url}  
\usepackage{csquotes}
\usepackage{natbib}

\usepackage[utf8]{inputenc}
\usepackage{url}

\usepackage{amsmath}
\usepackage{bbm}
\usepackage{bm}
\usepackage{cancel}
\usepackage{mathtools}

\usepackage{ascmac}  
\usepackage{algorithm}
\usepackage{algpseudocode}

\usepackage{fancyhdr}
\usepackage{tcolorbox}
\usepackage{tikz}

\usepackage{booktabs}
\usepackage{comment}
\usepackage{lastpage}
\usepackage{listings}
\usepackage{multibib}
\usepackage{multirow}
\usepackage[super]{nth}

\tcbuselibrary{breakable, skins, theorems}

\allowdisplaybreaks

\algtext*{EndFor}
\algtext*{EndWhile}
\algtext*{EndIf}
\algtext*{EndProcedure}
\algtext*{EndFunction}
\algnewcommand{\LineComment}[1]{\State \(\triangleright\) #1}

\newcites{appx}{References}

\newif\ifunderreview
\newif\ifsubmission
\newif\ifappendix

\underreviewfalse

\submissiontrue

\appendixtrue

\ifsubmission
\newcommand{\todo}[1]{}
\newcommand{\replace}[2]{}
\newcommand{\sw}[1]{}

\else
\newcommand{\todo}[1]{\textbf{\textcolor{red}{[TODO: #1]}}}

\newcommand{\replace}[2]{\textbf{\textcolor{red}{[del: \cancel{#1}]}}\textbf{\textcolor{blue}{[new: #2]}}}

\newcommand{\sw}[1]{\textbf{\textcolor{cyan}{[SW: #1]}}}

\usepackage[inline]{showlabels}

\fi

\makeatletter
\newcommand{\customlabel}[2]{%
   \protected@write \@auxout {}{\string \newlabel {#1}{{#2}{\thepage}{#2}{#1}{}} }%
   \hypertarget{#1}{}
}
\makeatother

\newcommand{\argmin}{\mathop{\mathrm{argmin}}}
\newcommand{\argmax}{\mathop{\mathrm{argmax}}}

\newcommand{\xv}{\boldsymbol{x}}
\newcommand{\X}{\mathcal{X}}

\newcommand{\D}{\mathcal{D}}

\newcommand{\secref}[1]{Section~\ref{#1}}
\renewcommand{\eqref}[1]{Eq.~(\ref{#1})}
\newcommand{\figref}[1]{Figure~\ref{#1}}

\DeclareFixedFont{\ttb}{T1}{txtt}{bx}{n}{12} 
\DeclareFixedFont{\ttm}{T1}{txtt}{m}{n}{12}  

\definecolor{deepblue}{rgb}{0,0,0.5}
\definecolor{deepred}{rgb}{0.6,0,0}
\definecolor{deepgreen}{rgb}{0,0.5,0}

\begin{document}
\title{
  Tree-Structured Parzen Estimator Can Solve Black-Box Combinatorial Optimization More Efficiently
}

\ifunderreview
\else
\author[1]{Kenshin Abe,}
\author[1]{Yunzhuo Wang,}
\author[1]{Shuhei Watanabe}
\affil[$\dagger$]{\texttt{\{kenshin,yunzhuo,shuheiwatanabe\}@preferred.jp}}
\affil[1]{Preferred Networks Inc., Japan}
\fi

\maketitle

\begin{abstract}
  Tree-structured Parzen estimator (TPE) is a versatile hyperparameter optimization (HPO) method supported by popular HPO tools.
  Since these HPO tools have been developed in line with the trend of deep learning (DL), the problem setups often used in the DL domain have been discussed for TPE such as multi-objective optimization and multi-fidelity optimization.
  However, the practical applications of HPO are not limited to DL, and black-box combinatorial optimization is actively utilized in some domains, e.g. chemistry and biology.
  As combinatorial optimization has been an untouched, yet very important, topic in TPE, we propose an efficient combinatorial optimization algorithm for TPE.
  In this paper, we first generalize the categorical kernel with the numerical kernel in TPE, enabling us to introduce a distance structure to the categorical kernel.
  Then we discuss modifications for the newly developed kernel to handle a large combinatorial search space.
  These modifications reduce the time complexity of the kernel calculation with respect to the size of a combinatorial search space.
  In the experiments using synthetic problems, we verified that our proposed method identifies better solutions with fewer evaluations than the original TPE.
  \ifunderreview
  \else
  Our algorithm is available in Optuna, an open-source framework for HPO.
  \fi
\end{abstract}

\section{Introduction}
The recent advance in deep learning and its hyperparameter (HP) sensitivity highlighted the importance of HP optimization (HPO)~\citep{henderson2018deep,vallati2021importance,zhang2021importance,sukthanker2022importance,wagner2022importance,yang2022tensor} such as Bayesian optimization~\citep{snoek2012practical,shahriari2015taking,garnett2023bayesian} and CMA-ES~\citep{hansen2016cma,loshchilov2016cma}.
To further make them accessible to end users, several open-source softwares such as Hyperopt~\citep{Bergstra_2015}, RayTune~\citep{liaw2018tune}, and Optuna~\citep{akiba2019optuna} have been developed so far.
These softwares support tree-structured Parzen estimator (TPE)~\citep{bergstra2011algorithms, watanabe2023tree}, a versatile HPO method, as their core algorithm.
Indeed, TPE has already played a pivotal role in improving the performance of real-world applications~\citep{happy-whale-and-dolphin,patton2023deep}.

However, HPO applications are not limited to deep learning as surveyed by \cite{li2024multi} and some domains consider black-box combinatorial optimization such as chemistry~\citep{wu2020bayesian} and biology~\citep*{khan2023toward}.
Although TPE has seen diverse enhancements such as multi-objective optimization~\citep{ozaki2020multiobjective,ozaki2022multiobjective}, constrained optimization~\citep{watanabe2022c,watanabe2023c}, meta-learning~\citep{watanabe2022multi,watanabe2023speeding}, and multi-fidelity optimization~\citep{falkner2018bohb}, the study on the combinatorial optimization using TPE has not been conducted and TPE still considers each combination to be equally similar, making TPE very inefficient in combinatorial optimization.

To mitigate the issue, we propose an efficient combinatorial algorithm for TPE that takes into account a user-defined distance structure between each category such as Hamming distance.
We first theoretically generalize the categorical kernel in TPE with the numerical kernel so that a distance structure can be naturally introduced to the categorical kernel.
Then we discuss two modifications for large combinatorial search spaces: (1) that enhances the runtime order and (2) that alleviates the overexploration during optimization.
In the experiments on synthetic problems, we demonstrate that our proposed method improves the performance from the original TPE and our modification enhances the sample efficiency in large combinatorial search spaces.
Our experiment code is available at 
\ifunderreview
\else
\url{https://github.com/nabenabe0928/optuna/tree/freeze/metric-tpe-experiments/}.
\fi

\section{Background \& Related Work}
In this section, we first define our problem setup and then explain the algorithm of TPE.

\subsection{Black-Box Combinatorial Optimization}
Black-box optimization aims to find the following optimizer:
\begin{equation}
\begin{aligned}
  \xv_{\mathrm{opt}} \in \argmin_{\xv \in \X} f(\xv)
\end{aligned}
\end{equation}
where $\X \coloneqq \X_1 \times \dots \times \X_D$ is the search space, $\X_d \subseteq \mathbb{R}$ for $d \in [D] \coloneqq \{1,\dots,D\}$ is the domain of the $d$-th HP, and $f: \X \rightarrow \mathbb{R}$ is an objective function.
Meanwhile, in black-box \emph{combinatorial} optimization, some dimensions $\X_d$ take a combinatorial domain, which is not limited to a subset of real numbers.
Typical examples of a combinatorial domain are permutations $\mathcal{S}_p$ of $p \in \mathbb{Z}_+$ and a $K$-ary number $[K]^L$ with a fixed length $L \in \mathbb{Z}_+$.
Although combinatorial optimization opens up a further application opportunity such as in chemistry and biology domains and there have been several works for combinatorial Bayesian optimization~\citep{baptista2018bayesian,oh2019combinatorial,wu2020bayesian,deshwal2023bayesian,papenmeier2023bounce}, efficient combinatorial optimization for TPE has not been discussed so far.

\subsection{Tree-Structured Parzen Estimator (TPE)}
TPE is a prevalent Bayesian optimization method used in many open-source softwares.
Given a set of observations $\D \coloneqq \{(\xv_n, f(\xv_n))\}_{n=1}^{N}$, TPE first splits the observations into two sets $\D_{\mathrm{good}} \coloneqq \{(x,f) \in \D | f < f^\gamma\}$ and $\D_{\mathrm{bad}} \coloneqq \D \setminus \D_{\mathrm{good}}$ where $f^{\gamma}$ is the $\gamma$ quantile in $\{f(\xv_n)\}_{n=1}^N$ and $\gamma \in [0, 1]$ is determined by a heuristic.
Then, we build the following two Parzen estimators:
\begin{equation}
\begin{aligned}
  p(\xv | \D_{\mathrm{good}}) = \frac{1}{|\D_{\mathrm{good}}|}\sum_{(\xv^\prime, \cdot) \in \D_{\mathrm{good}}}\frac{k(\xv, \xv^\prime)}{Z(\xv^\prime)},~~ p(\xv | \D_{\mathrm{bad}}) = \frac{1}{|\D_{\mathrm{bad}}|}\sum_{(\xv^\prime, \cdot) \in \D_{\mathrm{bad}}}\frac{k(\xv, \xv^\prime)}{Z(\xv^\prime)}, \\
\end{aligned}
\label{main:background:eq:two-parzen-estimators}
\end{equation}
where $Z(\xv^\prime) \coloneqq \int_{\xv \in \X} k(\xv, \xv^\prime)d\xv$ is a normalization factor and the kernel function $k$ is a product of the kernels in each dimension $k(\xv, \xv^\prime) = \prod_{d=1}^D k_d(x_d, x_d^\prime)$.
Gaussian kernel
\begin{equation}
\begin{aligned}
  k_d(x_d, x^\prime_d) = \exp\biggl(-\frac{1}{2}\biggl(\frac{x_d - x_d^\prime}{h}\biggr)^2\biggr)
\end{aligned}
\label{main:background:eq:numerical-kernel}
\end{equation}
is employed for numerical dimensions, and Aitchison-Aitken kernel
\begin{equation}
\begin{aligned}
  k_d(x_d, x^\prime_d) = \biggl(
    \frac{h}{C_d - 1}
  \biggr)^{1 - \delta(x_d, x_d^\prime)}
\end{aligned}
\label{main:background:eq:categorical-kernel}
\end{equation}
is employed for categorical dimensions.
Note that bandwidth $h$ is determined by a heuristic as discussed by \cite{watanabe2023tree}, $\delta: \mathbb{Z} \times \mathbb{Z} \rightarrow \{0,1\}$ is the Kronecker's delta function, and $C_d$ is the number of categories for the categorical parameter $x_d$.
The next HP is picked by the density ratio:
\begin{equation}
\begin{aligned}
  \xv_{N+1} \in \argmax_{\xv \in \mathcal{S}} \frac{p(\xv | \D_{\mathrm{good}})}{p(\xv | \D_{\mathrm{bad}})}
\end{aligned}
\end{equation}
where $\mathcal{S} \subseteq \X$ is a set of candidates sampled from $p(\xv | \D_{\mathrm{good}})$.
Since the combinatorial structure is currently handled by the Aitchison-Aitken kernel~\citep{aitchison1976multivariate}, which considers all choices equally similar, the optimization becomes very inefficient.
In this paper, we would like to mitigate this problem by deploying a user-defined distance metric between each combination.

\section{Efficient Combinatorial Optimization for Tree-Structured Parzen Estimator}

As discussed earlier, the problem of combinatorial optimization by TPE comes from the fact that TPE handles each category in a categorical parameter equally.
In this section, we first theoretically bridge the gap between the categorical and the numerical kernels, and discuss how we naturally introduce a user-defined distance metric.
Based on the formulation, we will propose a more efficient combinatorial algorithm for TPE.
As a na\"ive version of our algorithm requires the quadratic time complexity to the number of combinations $C_d$ at each sample phase, we revise the algorithm to improve the time complexity.

\subsection{Generalization of Categorical Kernel with Numerical Kernel}

To address this problem, we first would like to consider the generalization of the categorical kernel with the numerical kernel.
Transforming \eqref{main:background:eq:categorical-kernel}, we obtain the categorical kernel as the Gauss kernel-like form:
\begin{equation}
  \begin{aligned}
    \biggl(
    \frac{h}{C_d - 1}
    \biggr)^{1 - \delta(x_d, x_d^\prime)}
    = \biggl(
    \frac{h}{C_d - 1}
    \biggr)^{(1 - \delta(x_d, x_d^\prime))^2}
    =
    \exp\biggl(
    -\log\frac{C_d - 1}{h}(1 - \delta(x_d, x_d^\prime))^2
    \biggr).
  \end{aligned}
\end{equation}
Since $1 - \delta(x_d, x_d^\prime)$ is a distance metric, we can naturally replace $1 - \delta(x_d, x_d^\prime)$ with a user-defined metric $M_d: \X_d \times \X_d \rightarrow \mathbb{R}_{\geq 0}$ as follows:
\begin{equation}
  \begin{aligned}
    k_d(x_d, x_d^\prime) = \exp\biggl(
    -\frac{1}{2}\biggl(
      \frac{M_d(x_d, x_d^\prime)}{\beta}
      \biggr)^2
    \biggr)~\mathrm{where}~\beta = \frac{M_d^{\max}}{\sqrt{2 \log\frac{C_d - 1}{h}}},\mathrm{~and~}
  \end{aligned}
  \label{main:methods:eq:combinatorial-kernel}
\end{equation}
$M_d^{\max} \coloneqq \max_{[x_d^{(1)}, x_d^{(2)}] \in \X_d \times \X_d} M_d(x_d^{(1)},x_d^{(2)})$.
Note that $\beta$ above is determined so that \eqref{main:methods:eq:combinatorial-kernel} falls back to \eqref{main:background:eq:categorical-kernel} when $M_d(x_d, x_d^\prime) = M_d^{\max}$ holds.
Our new formulation enables a natural extension of a categorical parameter to a numerical parameter given an appropriate distance metric.

\subsection{Practical Considerations}

In the previous section, we have discussed the theoretical formulation of the extension of TPE to the combinatorial setup.
Although this formulation naturally bridges the gap between categorical and numerical kernels, it misses out practical considerations in terms of computational efficiency and performance for problem setups where there are many combinations.

\subsubsection{Efficient Approximation of Maximum Distance}
In our proposition, the bottlenecks lie in the computation of $M_d^{\max}$, which requires the time complexity of $\Theta(C_d^2)$.
Instead of this calculation, we use $M_d^{\max, \star}(x_d^\prime) \coloneqq \max_{x_d^{(1)} \in \X_d} M_d(x_d^{(1)}, x_d^\prime)$ in $\beta$ of \eqref{main:methods:eq:combinatorial-kernel} for each basis in \eqref{main:background:eq:two-parzen-estimators}.
By doing so, the time complexity of the $\beta$ calculation reduces to $\Theta(C_d \min(C_d, N^{\mathrm{unique}}_{d}))$ where $N^{\mathrm{unique}}_{d}$ is the unique count of the $d$-th HP value in a given set of observations.
The time reduction effect is significant especially when the number of HP evaluations $N$ is much smaller than $C_d$.
Note that the maximum error we might yield is $M_d^{\max, \star}(x_d^\prime) \leq M_d^{\max} \leq 2 M_d^{\max, \star}(x_d^\prime)$
due to the triangle inequality, i.e. $M_d(x_d^{(1)}, x_d^{(2)}) \leq M_d(x_d^{(1)}, x_d^\prime) + M_d(x_d^\prime, x_d^{(2)})$.

\subsubsection{Modification of Combinatorial Kernel}
\label{main:section:modification-of-combinatorial-kernel}
In this section, we consistently use $h = \frac{C_d-1}{|\D_{\cdot}| + 1} \coloneqq \frac{C_d-1}{N_{\cdot} + 1}$ used in the Optuna implementation for practical considerations.
In practice, we may encode several categorical parameters into a combinatorial parameter, i.e. a categorical parameter with a distance metric.
For example, we might want to include a categorical parameter $x \in [2^K - 1]$ with the Hamming distance $M_{\mathrm{Ham}}$, leading to $(\frac{1}{N_{\cdot} + 1})^{\frac{M_{\mathrm{Ham}}(x, x^\prime)}{K}}$ in our proposition based on \eqref{main:methods:eq:combinatorial-kernel}.
Notice that the maximum distance of $M_{\mathrm{Ham}}$ on $[2^K - 1]$ is $K$, and thus the maximum possible exponent is $1$.
Meanwhile, this categorical parameter can also be represented as $K$ binary parameters $[B_1, \dots, B_K]\in \{0,1\}^K$, leading to $\prod_{i=1}^K (\frac{1}{N_{\cdot} + 1})^{1 - \delta(B_i, B_i^\prime)} = (\frac{1}{N_{\cdot} + 1})^{K - \sum_{i=1}^K \delta(B_i, B_i^\prime)}$ in the original TPE formulation based on \eqref{main:background:eq:categorical-kernel}.
If we define $x = \sum_{i=0}^{K - 1} B_i 2^i$ and $x^\prime = \sum_{i=0}^{K - 1} B_i^\prime 2^i$, both representations handle the same search space.

However, the maximum exponent of our proposition is $1$ and that of the original formulation is $K$, yielding oversmoothing in our kernel and leaning to too much exploration during optimization.
The oversmoothing poses bad performance especially for a large combinatorial search space due to limited coverage of the search space.  
To alleviate this problem, we modify $\beta$ in \eqref{main:methods:eq:combinatorial-kernel} as follows:
\begin{equation}
  \begin{aligned}
    \beta^\prime = \frac{\beta}{\sqrt{\log_{b_d}C_d}} = \frac{M_d^{\max}}{\sqrt{2 \log \frac{C_d - 1}{h}\log_{b_d}C_d}}
  \end{aligned}
  \label{main:methods:eq:modification-combinatorial-kernel}
\end{equation}
where $b_d \in \mathbb{R}_+$ is the control parameter in our algorithm.
This formulation assumes that $\log_{b_d}C_d$ categorical parameters are encoded into a combinatorial parameter $x_d$ and a smaller $b_d$ leads to more exploitation.
Although users can customize the control parameter $b_d$, e.g. $b_d$ should be $2$ in the example above, we fix $b_d = 6$ as a default value.
Note that we performed the ablation study of $b_d$ in Appendix~\ref{appx:section:ablation-study}.

\section{Experiments}
\label{main:section:experiments}
In this section, we verify that our proposition increases the sample efficiency of TPE on combinatorial optimization using the following two synthetic problems:
\begin{enumerate}
  \vspace{-1mm}
  \item \texttt{EmbeddingCosine}: minimizes the cosine similarity $f(\xv | \xv_{\mathrm{opt}}) = 1 - \frac{\xv \cdot \xv_{\mathrm{opt}}}{\|\xv\| \|\xv_{\mathrm{opt}}\|}$ given the optimal $\xv_{\mathrm{opt}} \in \X \subseteq [0, 1]^K$ with a distance metric $M(\xv, \xv^\prime) = 1 - \frac{\xv \cdot \xv^\prime}{\|\xv\| \|\xv^\prime\|}$, and
  \vspace{-1mm}
  \item \texttt{PermutationShiftL1}: minimizes the L1 norm $f(\xv | \xv_{\mathrm{opt}}) = \|\boldsymbol{s} - \boldsymbol{s}_{\mathrm{opt}} + (a - a_{\mathrm{opt}}) \boldsymbol{1}_p \|_1$ given the optimal $\xv_{\mathrm{opt}} = \{\boldsymbol{s}_{\mathrm{opt}}, a_{\mathrm{opt}}\}\in \X = \mathcal{S}_p \times [-p, p]$ with a distance metric $M(\boldsymbol{s}, \boldsymbol{s}^\prime) = \|\boldsymbol{s} - \boldsymbol{s}^\prime\|_1$ where $p$ is a positive integer, $\mathcal{S}_p$ is a set of all the permutations of $[p]$, and $\boldsymbol{1}_p$ is a $p$-dimensional vector having only $1$ as elements. 
  \vspace{-1mm}
\end{enumerate}
For both problems, we first fix $\xv_{\mathrm{opt}}$ and each optimizer aims to find $\xv_{\mathrm{opt}}$ during optimization.
\texttt{EmbeddingCosine} first generates $\X$ by uniformly sampling $C$ elements from $[0, 1]^K$ with replacement.
The control parameters of \texttt{EmbeddingCosine} are $C$ and $K$ and those of \texttt{PermutationShiftL1} are $p$.
In our experiments, we used $[C, K] \in \{[500, 8], [1000, 16]\}$ and $p \in \{6, 7\}$.
Note that there are $6! = 720$ and $7! = 5040$ possible combinations respectively in the combinatorial domains of \texttt{PermutationShiftL1} with $p \in \{6, 7\}$.
We used the original TPE in Optuna and random search as baselines, and used the proposed method without the modification discussed in \secref{main:section:modification-of-combinatorial-kernel}.
In our experiments, we consistently used the default setup of \texttt{TPESampler} except for \texttt{multivariate=True} and evaluated $100$ HPs for each optimization.
All the setups were run over $10$ different random seeds.
The whole experiment took 2 CPU hours on a machine with Intel Core i7-1255U.
Our experiment code is available at 
\ifunderreview
\else
\url{https://github.com/nabenabe0928/optuna/tree/freeze/metric-tpe-experiments/}.
\fi

\begin{figure}
  \centering
  \includegraphics[width=0.95\textwidth]{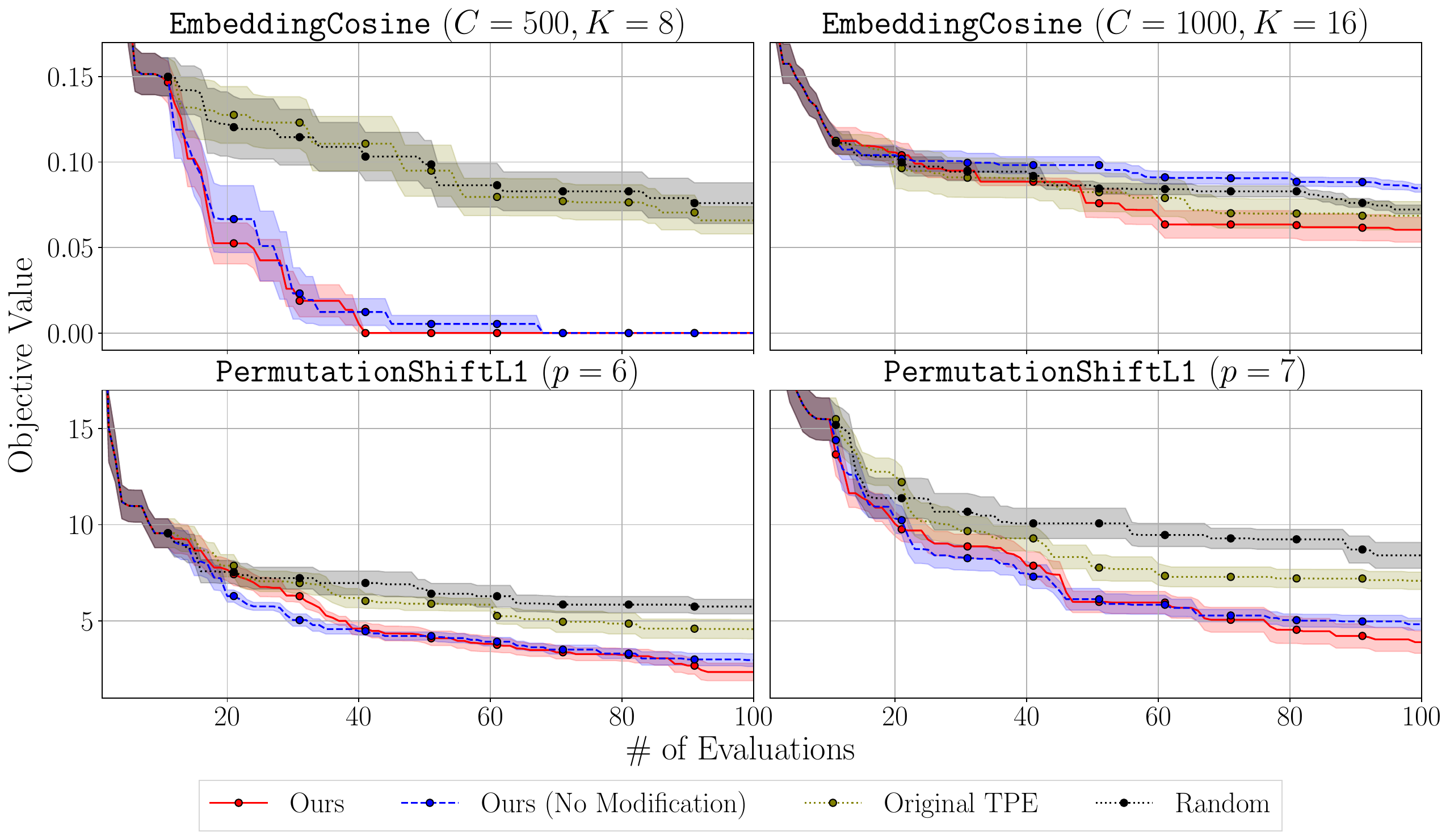}
  \vspace*{-1mm}
  \caption{
    Performance over time of each method.
    Each line and weak-color band show the mean and the standard error over 10 random seeds.
    The $x$- and $y$-axes represent the number of HP evaluations and the best objective value.
    \textbf{Top}: the results on \texttt{EmbeddingCosine} ($C=500, K=8$ (\textbf{Left}), and $C=1000, K=16$ (\textbf{Right})).
    \textbf{Bottom}: the results on \texttt{PermutationShiftL1} ($p=6$ (\textbf{Left}), and $p=7$ (\textbf{Right})).
  }
  \vspace*{-1mm}
  \label{main:experiments:fig:perf-over-time}
\end{figure}

\figref{main:experiments:fig:perf-over-time} shows the results of the experiments.
As can be seen, the performance of the original TPE and random search is not distinguishable except for \texttt{PermutationShiftL1} with $p = 7$.
On the other hand, the weak-color bands for the proposed method and the original TPE do not overlap, implying that the proposed method was superior to the original TPE on the provided large combinatorial problems.
Furthermore, the modification in \eqref{main:methods:eq:combinatorial-kernel} improved the performance especially for the ones with larger combinatorial search spaces (\textbf{Right} of each row) as expected.
More importantly, the modification did not degrade the performance of our proposed method on all the setups.
Looking at \textbf{Top Right} of \figref{main:experiments:fig:perf-over-time}, we can observe that overexploration even made the performance worse than random search without the modification.
Although more diverse experiments are required to draw solid conclusions, our modification made TPE more robust in our experiments.

\section{Broader Impact \& Limitations}
\label{main:section:limitations-and-broader-impact}
\ifunderreview
\else
On the one hand, as TPE is used in very popular HPO frameworks and our proposed method is available in Optuna, our proposition will reduce the time necessary before users start a project involving a black-box combinatorial optimization problem.
\fi
On the other hand, although our method can solve combinatorial optimization more efficiently compared to the original TPE, both TPE and our approach suffer from memory and time complexities when applied to a large combinatorial search space because each kernel in Parzen estimators must be stored in memory, which requires $\Omega(C_d)$.
Therefore, it is not possible to apply our method to an objective function with a huge combinatorial search space such as permutations of 15+ elements.
One possible solution would be to limit the combinations to search at each sample step by sampling them from the original search space.
Another limitation of our paper is the inadequate methodological evaluation to claim a more general high performance.
Extensive benchmarking on more practical problems, e.g. those provided by \cite{dreczkowski2024framework}, and scalability experiments using parallel setups~\citep{watanabe2023python,watanabe2024fast}, is our future work.

\section{Conclusion}
\label{main:section:conclusion}
We proposed a formulation to fill the gap between categorical and numerical kernels in TPE, leading to more efficient combinatorial optimization by TPE and allowing users to introduce a distance structure for categorical parameters.
Furthermore, we modified our algorithm to enhance the original quadratic time complexity in the kernel calculation and alleviate an overexploration issue for a large combinatorial search space.
In the experiments, we demonstrated our proposed method effectively improved the performance and was not degraded for large combinatorial search spaces.

\newpage
\ifunderreview
\else
\section*{Acknowledgement}
We thank Yoshihiko Ozaki for the paper review and detailed comments.
We also appreciate AutoML team members in Preferred Network Inc. for insightful feedback from the early stages of the method development.
\fi

\bibliographystyle{apalike}
\bibliography{ref}

\newpage

\ifappendix
\clearpage
\appendix
\section*{Appendix}

\section{Ablation Study of $b_d$}
\label{appx:section:ablation-study}

As discussed in \secref{main:section:modification-of-combinatorial-kernel}, a smaller $b_d$ leads to more exploitation and a larger $b_d$ leads to more exploration, so we would like to see the effect of this parameter to the optimization performance. 
\figref{appx:experiments:fig:ablation-study} shows the ablation study of $b_d \in \{2, 3, \dots, 10\}$ in \eqref{main:methods:eq:modification-combinatorial-kernel}.
For \texttt{PermutationShiftL1} and \texttt{EmbeddingCosine} with $C=500, K=8$, it is hard to distinguish the performance between each $b_d$.
On the other hand, $b_d$ has a relatively larger impact in \texttt{EmbeddingCosine} with $C=1000, K=16$ than the others.
This is probably due to its search space size.
Theoretically speaking, the search space of \texttt{EmbeddingCosine} with $C=1000, K=16$ is a $17$-dimensional continuous space although $C$ is $1000$ in this setup.
Since the theoretical search space size is much bigger than the other setups, the optimization would also be the hardest.
As the optimization budget, i.e. $100$ HP evaluations, was severely limited in our setup, the exploitation-leaning search, i.e. with a lower $b_d$, probably outperformed the exploration-leaning search, i.e. with a higher $b_d$. 

\begin{figure}
  \centering
  \includegraphics[width=0.98\textwidth]{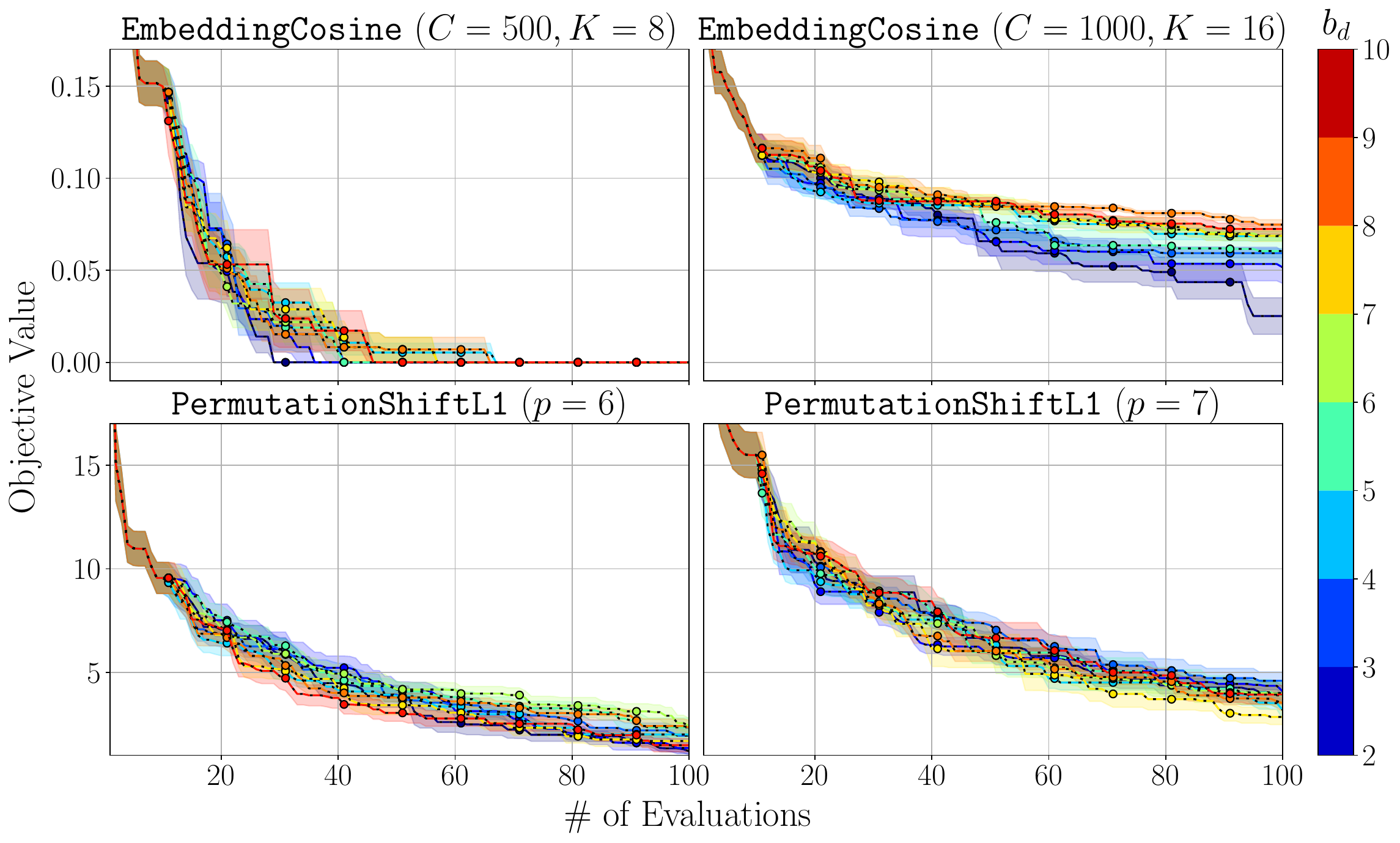}
  \caption{
    Ablation study of $b_d$ in our method.
    Each line and weak-color band show the mean and the standard error over 10 random seeds.
    $b_d$ becomes higher as the plot color approaches red.
    The $x$- and $y$-axes represent the number of HP evaluations and the best objective value.
    \textbf{Top}: the results on \texttt{EmbeddingCosine} ($C=500, K=8$ (\textbf{Left}), and $C=1000, K=16$ (\textbf{Right})).
    \textbf{Bottom}: the results on \texttt{PermutationShiftL1} ($p=6$ (\textbf{Left}), and $p=7$ (\textbf{Right})).
  }
  \label{appx:experiments:fig:ablation-study}
\end{figure}

\else

\customlabel{data1}{1}
\customlabel{data2}{2}

\fi

\end{document}